\documentclass{article} 
\usepackage{nips11submit_e,times}
\usepackage{latexsym}
\usepackage{amsmath}
\usepackage{url}
\usepackage{graphicx}
\usepackage{multirow}

\usepackage{wrapfig,comment}

\usepackage{lastpage,graphics,graphicx}
\usepackage{amsmath,amssymb,amsfonts}
\usepackage{url}

\usepackage{enumitem}
   {
      \begin{enumerate}[nolistsep,label=\Alph*.,ref=\Alph{enumi},
        leftmargin=0pt,labelsep=15pt,align=left,
        labelwidth=0.1cm,itemindent=1.0cm]}
  {\end{enumerate}}
\usepackage{enumitem}
   {
      \begin{enumerate}[nolistsep,label=\Alph*.,ref=\Alph{enumi},
        leftmargin=0pt,labelsep=15pt,align=left,
        labelwidth=0.1cm,itemindent=1.0cm]}
  {\end{enumerate}}
   {
      \begin{enumerate}[nolistsep,label=\Alph*.,ref=\Alph{enumi},
        leftmargin=0pt,labelsep=15pt,align=left,
        labelwidth=1.5cm,itemindent=2.0cm]}
  {\end{enumerate}}

\title{Learning New Facts From Knowledge Bases With \\ Neural Tensor Networks and Semantic Word Vectors}

\author{\hspace{-0.3cm}Danqi Chen, Richard Socher, Christopher D. Manning, Andrew Y. Ng\\
Computer Science Department, Stanford University, Stanford, CA 94305, USA\\
{\tt \small \{danqi,manning,ang\}@stanford.edu, richard@socher.org} \\
}

\nipsfinalcopy 

\begin{document}

\maketitle

\begin{abstract}
Knowledge bases provide applications with the benefit of easily
accessible, systematic relational knowledge but often suffer in
practice from their incompleteness and lack of knowledge of new
entities and relations.
Much work has focused on building or extending them by finding
patterns in large unannotated text corpora. In contrast, here we mainly aim to
complete a knowledge base by predicting additional true relationships
between entities, based on generalizations that can be
discerned in the given knowledgebase.
We introduce a neural tensor network (NTN) model which predicts new
relationship entries that can be added to the
database. This model can be improved by initializing entity
representations with word vectors learned in 
an unsupervised fashion from text, and when doing this, existing
relations can even be queried for
entities that were not present in the database. 
Our model generalizes and outperforms existing models for this
problem, and can classify unseen relationships in WordNet with an accuracy of 75.8\%.
\end{abstract}

\section{Introduction}
Ontologies and knowledge bases such as WordNet \cite{Miller1995} or Yago \cite{yago} are extremely useful resources for query expansion \cite{Graupmann}, coreference resolution \cite{ngCardie2002}, question answering (Siri), information retrieval (Google Knowledge Graph), or generally providing inference over structured knowledge to users. Much work has focused on extending existing knowledge bases \cite{snowWordnet,ReVerb2011,yago} using patterns or classifiers applied to large corpora.

We introduce a model that can accurately learn to add additional facts to a database using only that database.
This is achieved by representing each entity (i.e., each object or individual) in the database by a vector that can capture facts and their certainty about that entity. Each relation is defined by the parameters of a novel neural tensor network which can explicitly relate two entity vectors and is more powerful than a standard neural network layer.

Furthermore, our model allows us to ask whether even entities that were not in the database are in certain relationships by simply using distributional word vectors. These vectors are learned by a neural network model \cite{Turian2010} using unsupervised text corpora such as Wikipedia. They capture syntactic and semantic information and allow us to extend the database without any manually designed rules or additional parsing of other textual resources.

The model outperforms previously introduced related models such as that of Bordes et al. \cite{Bordes2012}. We evaluate on a heldout set of relationships in WordNet. The accuracy for predicting unseen relations is 75.8\%. We also evaluate in terms of ranking. For WordNet, there are 38,696 different entities and we use 11 relationship types. On average for each left entity there are 100 correct entities in a specific relationship. For instance, \emph{dog} has many hundreds of hyponyms such as \emph{puppy, barker} or \emph{dachshund}. In 20.9\% of the relationship triplets, the model ranks the correct test entity in the top 100 out of 38,696 possible entities.

\section{Related Work}
There is a vast amount of work extending knowledge bases using external corpora \cite{snowWordnet,ReVerb2011,yago}, among many others. In contrast, little work has been done in extensions based purely on the knowledge base itself. The work closest to ours is that by Bordes et al. \cite{bordes2011}. We implement their approach and compare to it directly. Our model outperforms it by a significant margin in terms of both accuracy and ranking. Both models can benefit from initialization with unsupervised word vectors.

Another related approach is that by Sutskever et al. \cite{ilya2009} who use tensor factorization and Bayesian clustering for learning relational structures. Instead of clustering the entities in a nonparametric Bayesian framework we rely purely on learned entity vectors. Their computation of the truth of a relation can be seen as a special case of our proposed model. Instead of using MCMC for inference, we use standard backpropagation which is modified for the Neural Tensor Network. Lastly, we do not require multiple embeddings for each entity. Instead, we consider the subunits (space separated words) of entity names. This allows more statistical strength to be shared among entities.

Many methods that use knowledge bases as features such as \cite{Graupmann,ngCardie2002} could benefit from a method that maps the provided information into vector representations. We learn to modify unsupervised word representations via grounding in world knowledge. This essentially allows us to analyze word embeddings and query them for specific relations. Furthermore, the resulting vectors could be used in other tasks such as NER \cite{Turian2010} or relation classification in natural language \cite{SocherEtAl2012:MVRNN}.

Lastly, Ranzato et al. \cite{3wayRBM} introduced a factored 3-way Restricted Boltzmann Machine which is also parameterized by a tensor.

\section{Neural Tensor Networks}
In this section we describe the full neural tensor network. We begin by describing the representation of entities and continue with the model that learns entity relationships.

We compare using both randomly initialized word vectors and pre-trained $100$-dimensional word vectors from the unsupervised model of Collobert and Weston \cite{collobert2008:deep,Turian2010}. Using free Wikipedia text, this model learns word vectors by predicting how likely it is for each word to occur in its context. The model uses both local context in the window around each word and global document context.
Similar to other local co-occurrence based vector space models, the resulting word vectors capture distributional syntactic and semantic information. For further details and evaluations of these embeddings, see \cite{Bengio2003,collobert2008:deep,Huang2012}.

For cases where the entity name has multiple words, we simply average the word vectors.

The Neural Tensor Network (NTN) replaces the standard linear layer with a bilinear layer that directly relates the two entity vectors. Let $e_1,e_2 \in \mathbb{R}^d$ be the vector representations of the two entities. We can compute a score of how plausible they are in a certain relationship $R$ by the following NTN-based function:
\begin{equation}
g(e_1, R, e_2) = U^T f\left(e_1^T W^{[1:k]}_R e_2 + V_R \left[ \begin{matrix} 
e_1\\
e_2\\
\end{matrix}
\right] +b_R \right),
\label{eq:NTN}
\end{equation}
where $f = \tanh$ is a standard nonlinearity. We define $W^{[1:k]} \in \mathbb{R}^{d \times d \times k}$ as a tensor and the bilinear tensor product results in a vector $h \in \mathbb{R}^k$, where each entry is computed by one slice of the tensor:
\begin{equation}
h_i = e_1^TW^{[i]}e_2.
\label{eq:tensorHid}
\end{equation}
The remaining parameters for relation $R$ are the standard form of a neural network: $V_R \in \mathbb{R}^{k \times 2d}$ and $ U \in \mathbb{R}^{k}, b_R \in \mathbb{R}^k$.

The main advantage of this model is that it can directly relate the two inputs instead of only implicitly through the nonlinearity.
The bilinear model for truth values in \cite{ilya2009} becomes a special case of this model with $V_R=\mathbf{0}, b_R = 0, k=1,f=identity$. 

In order to train the parameters $W, U, V, E, b$, we minimize the following contrastive max-margin objective:
\begin{equation}
J(W,U,V,E, b) = \sum_{i=1}^N \sum_{c=1}^C \max(0,1-g(e^{(i)}_1, R^{(i)}, e^{(i)}_2) + g(e^{(i)}_1, R^{(i)}, e_c)),
\label{eq:obj}
\end{equation}
where $N$ is the number of training triplets and we score the correct relation triplets higher than a corrupted one in which one of the entities was replaced with a random entity. For each correct triplet we sample $C$ random corrupted entities.

The model is trained by taking gradients with respect to the five sets of parameters and using minibatched L-BFGS.

\section{Experiments}
In our experiments, we follow the data settings of WordNet in \cite{bordes2011}. 
There are a total of 38,696 different entities and 11 relations. We use 112,581 triplets for training,  2,609 for the development set and 10,544 for final testing.

The WordNet relationships we consider are \emph{has instance, type of, member meronym, member holonym, part of, has part, subordinate instance of, domain region, synset domain region, similar to, domain topic}.

We compare our model with two models in Bordes et al.\ \cite{bordes2011, Bordes2012}, which have the same goal as ours. The model of \cite{bordes2011} has the following scoring function:
\begin{equation}
g(e_1, R, e_2) = \|W_{R,left}e_1 - W_{R,right} e_2\|_1,
\label{eq:bordes}
\end{equation}
where $W_{R,left},W_{R,right} \in \mathbb{R}^{d \times d}$. The model of \cite{Bordes2012} also maps each relation type to an embedding $e_R \in \mathbb{R}^d$ and scores the relationships by:
\begin{equation}
g(e_1, R, e_2) = - (W_1 e_1 \otimes W_{rel,1} e_R + b_1) \cdot (W_2 e_2 \otimes W_{rel,2} e_R + b_2),
\label{eq:bordes12}
\end{equation}
where $W_1, W_{rel,1}, W_2, W_{rel,2} \in \mathbb{R}^{d \times d}, b_1, b_2 \in \mathbb{R}^{d \times 1}$. In the comparisons below, we call these two models the \emph{similarity model} and the \emph{Hadamard model} respectively. While our function scores correct triplets highly, these two models score correct triplets lower. All models are trained in a contrastive max-margin objective functions.

Our goal is to predict ``correct'' relations $(e_1, R, e_2)$ in the testing data. We can compute a score for each triplet $(e_1, R, e_2)$.  We can consider either just a classification accuracy result as to whether the relation holds, or look at a ranking of $e_2$, for considering relative confidence in particular relations holding. We use a different evaluation set from Bordes et al.  \cite{bordes2011} because it has became apparent to us and them that there were issues of overlap between their training and testing sets which impacted the quality and interpretability of their evaluation.


\subsection*{Ranking}
For each triplet $(e_1, R, e_2)$, we compute the score $g(e_1, R, e)$ for all other entities in the knowledge base $e \in E$. We then sort values by decreasing order and report the rank of the correct entity $e_2$.

For WordNet the total number of entities is $|E| = 38,696$. Some of the questions relating to triplets are of the form ``A is a type of ?'' or ``A has instance ?'' Since these have multiple correct answers, we report the percentage of times that $e_2$ is ranked in the top $100$ of the list (recall @ 100). The higher this number, the more often the specific correct test entity has likely been correctly estimated.

After cross-validation of the hyperparameters of both models on the development fold, our neural tensor net obtains a ranking recall score of 20.9\% while the similarity model achieves 10.6\%, and the Hadamard model achieves only 7.4\%.  The best performance of the NTN with random initialization instead of the semantic vectors drops to 16.9\% and the similarity model and the Hadamard model only achieve 5.7\% and 7.1\%.

\subsection*{Classification}
In this experiment, we ask the model whether any arbitrary triplet of entities and relations is true or not. With the help of the large vocabulary of semantic word vectors, we can query whether certain WordNet relationships hold or not even for entities that were not originally in WordNet. 


We use the development fold to find a threshold $T_R$ for each relation such that if  $f(e_1, R, e_2) \geq T_R$, the relation $(e_1, R, e_2)$ holds, otherwise it is considered false. In order to create negative examples, we randomly switch entities and relations from correct testing triplets, resulting in a total of $2\times 10,544$ triplets. The final accuracy is based on how many of of triplets are classified correctly. 

The Neural Tensor Network achieves an accuracy of 75.8\% with semantically initialized entity vectors and 70.0\% with randomly initialized ones. In comparison, the similarity based model only achieve 66.7\% and 51.6\%, the Hadamard model achieve 71.9\% and 68.2\% with the same setup.
All models improve in performance if entities are represented as an average of their word vectors but we will leave experimentation with this setup to future work.

\section{Conclusion}
We introduced a new model based on Neural Tensor Networks. Unlike previous models for predicting relationships purely using entity representations in knowledge bases, our model allows direct interaction of entity vectors via a tensor. This architecture allows for much better performance in terms of both ranking correct answers out of tens of thousands of possible ones and predicting unseen relationships between entities.
It enables the extension of databases even without external textual resources but can also benefit from unsupervised large corpora even without manually designed extraction rules.


\small{
\bibliographystyle{unsrt}
\bibliography{bib}
}

\end{document}